\documentclass{Interspeech2024}
\usepackage{cleveref}

\interspeechcameraready

\title{Quantification of stylistic differences in human- and ASR-produced transcripts of African American English}

\name[affiliation={1}]{Annika}{Heuser}
\name[affiliation={2}]{Tyler}{Kendall}
\name[affiliation={3}]{Miguel}{del Rio}
\name[affiliation={3}]{Quinten}{McNamara}
\name[affiliation={3}]{Nishchal}{Bhandari}
\name[affiliation={3}]{Corey}{Miller}
\name[affiliation={3}]{Migüel}{Jetté}

\address{
  $^1$University of Pennsylvania, USA;
  $^2$University of Oregon, USA;
  $^3$Rev.com, USA}
\email{aheuser@sas.upenn.edu, tsk@uoregon.edu, miguel.delrio@rev.com}

\keywords{automatic speech recognition, evaluation, transcription variation, bias, African American English, AAE}

\begin{document}

\maketitle

\begin{abstract}
Common measures of accuracy used to assess the performance of automatic speech recognition (ASR) systems, as well as human transcribers, conflate multiple sources of error. Stylistic differences, such as verbatim vs non-verbatim, can play a significant role in ASR performance evaluation when differences exist between training and test datasets. The problem is compounded for speech from underrepresented varieties, where the speech to orthography mapping is not as standardized. We categorize the kinds of stylistic differences between 6 transcription versions, 4 human- and 2 ASR-produced, of 10 hours of African American English (AAE) speech. Focusing on verbatim features and AAE morphosyntactic features, we investigate the interactions of these categories with how well transcripts can be compared via word error rate (WER). The results, and overall analysis, help clarify how ASR outputs are a function of the decisions made by the training data’s human transcribers.
\end{abstract}

\section{Introduction} 

 Word error rate (WER) is the standard metric for automatic speech recognition (ASR) evaluation, widely used across industry and research. However, WER is readily affected by the properties of an ASR system’s training and test data. All the idiosyncrasies of the chosen reference transcript play a role in how the system is trained, refined, and assessed. 
 This means a system's performance could degrade if the reference doesn't reflect the training data's idiosyncrasies, many of which could be considered stylistic in nature.
 Transcription styles are not a novel idea: in fact, companies like Rev\footnote{\url{https://www.rev.com/blog/resources/verbatim-transcription}} produce distinct verbatim and non-verbatim transcripts. Verbatim transcription includes filler words, such as ``um" and ``uh," false starts, and interjections, while non-verbatim transcription allows for light editing, still preserving the content of what was said in the audio. 
 Both are valid styles, but directly comparing a verbatim vs non-verbatim transcript would misleadingly highlight several ``errors", even between humans. To add to the complications, transcription companies commonly assign separate chunks of a single long file to different transcribers in order to maintain reasonable delivery times.

Deep Learning models are well known to capture their training data's distribution. We posit that during training, models acquire additional types of stylistic proclivities which can be explicitly observed, even beyond the verbatim vs non-verbatim axis of variation. It is precisely our limited knowledge about these proclivities that makes comparing different ASR models more challenging. Within the WER evaluation paradigm, a model would be penalized for not having a stylistic proclivity that another model and the reference transcript share. It is, however, critical that we can accurately compare different ASR models, in order to determine which architectures, training strategies, etc. are  most effective.

We define transcription style as the collection of decisions made in contexts where there are multiple reasonable alternatives for how to transform the audio signal into an orthographic representation. Additionally, style can make a transcript better suited to its purpose. Not all the differences between two transcripts are stylistic in nature -- for example: some might be perceptual disagreements, while others could of course be actual errors, like typos. Some of the differences might seem stylistic to one person but not another, because they do not agree on whether a given transcription choice was among the set of reasonable alternatives for the given context. 
Bucholtz \cite[p.\ 1452]{bucholtz2000politics} describes a situation where transcribers might be trying to do the ``original speaker a favor by `cleaning up'" their speech. However, some choices could be considered counter-productive or harmful because they might misrepresent the speech of an individual or community. Consequently, many researchers would not want ASR systems to replicate these. 

The current project collects multiple transcripts of a subset of the sociolinguistic interviews contained in the Corpus of Regional African American Language (CORAAL) \cite{kendall2018corpus} and characterizes  differences between them. We collected multiple transcripts to demonstrate that professional transcripts of the same audio can differ substantially, with concomitant effect on WER. The original CORAAL transcripts were transcribed and corrected by multiple researchers familiar with AAE, but the other transcripts are still professional-grade. As an underrepresented variety of English \cite{lanehart2015oxford}, AAE's orthography is not nearly as conventionalized as that of Standard American English (SAE).
We divide our transcripts into two groups: those produced by humans and those produced by ASR systems. We compare the distributions of differences within and across these two groups by categorizing the types of differences.


In this paper, we examine three categories of transcription differences, which serve as hypotheses for the potential \textit{sources} of the differences. For example, a verbatim vs non-verbatim category or hypothesis posits that any given difference between two transcripts could be due to the transcribers having different verbatim objectives. We then quantify what percentage of the time this hypothesis is true for any transcript pairing. In addition to 1) the verbatim vs non-verbatim hypothesis, we also test 2) whether morpho-syntactic features that differentiate AAE from SAE and 3) whether different reduction and contraction orthographic representations (e.g. ``going to" vs ``gonna" and ``she will" vs ``she'll") account for the differences. 

The morpho-syntactic hypothesis allows us to investigate ASR bias against AAE, and potentially identify its source. The greater the percentage of transcript differences that are accounted for by the morpho-syntactic hypothesis, the more one transcript or the other might be transcribing AAE as SAE. By applying the same test to ASR output and human-produced transcripts from the same distribution as the training data, we can track the extent to which the ASR system is emulating the human rates of transcription decisions regarding AAE morpho-syntactic features. 

These 3 hypotheses serve as our metrics to quantify the stylistic differences across transcript versions. The 3 categories examined here are not meant to capture the full range of possible differences, but we hope they can contribute to a complete ontology of the axes of transcription variation, which is left for future work.

\section{Background} 


Inter-transcriber variation, while under-explored in the context of ASR, has been examined in allied fields, such as phonetics, conversation analysis, and forensic linguistics. The analogy to verbatim vs non-verbatim in phonetics is narrow vs broad transcription. As might be expected since broad uses fewer symbols/distinctions than narrow, \cite{shriberg1991reliability} found that inter-transcriber reliability was higher for broad transcription. Nonetheless, broad phonetic transcripts are still much ``narrower" than word-level ASR transcripts. Conversation analysts often focus closely on transcriber decisions and agreement, in ways that are relevant to the interests of this paper, but focus on a wider-range of speech phenomena (such as pauses and intonation) \cite{patterson1996preliminary}. Forensic linguists are often concerned with content agreement between humans and also ASR transcriptions \cite{loakes2022does}. Given the stakes of transcription in legal contexts, it is perhaps unsurprising that forensic linguists have considered categories of transcription differences. For instance, \cite{love2021specifying} generated a difference ontology by manually examining eight transcripts of an audio recording produced by different linguistically trained transcribers. It consisted of 1) omitted/additional speech, 2) splitting of turns, 3) phonetic similarity, and 4) lexical variation. 1) corresponds to the verbatim vs non-verbatim distinction and 2) corresponds to speaker attribution/diarization differences as opposed to word-level phenomena. Like 1), 3) is a hypothesis of the origin of transcription differences, namely that they are the result of perceptual differences. Finally, 4)  corresponds to the rest of the transcription differences. 

Recent work, in both forensic linguistics and in ASR research, has investigated transcription accuracy on non-standard varieties of English \cite{del2023accents}, particularly on AAE \cite{jones2019testifying,koenecke2020racial,wassink2022uneven}. However, work thus far has not investigated the categories underlying disagreements and inaccuracies across human- and ASR-produced transcripts of the same audio data. 




\section{Data}
We selected 27 files from CORAAL, corresponding to about 10 hours of audio. For each file we produced 6 transcript versions (referred to simply as versions from this point forward).
\\ \\
\textbf{\textit{Human Versions}} 
\begin{itemize}
  \item \textbf{CORAAL}: The original transcript from the CORAAL corpus, produced by \cite{kendall2018corpus}.
  \item \textbf{Rev}: Generated by soliciting a verbatim transcript through the web interface of Rev.com. 
  \item \textbf{Rev (+AA tag)}: Generated exactly like the \textbf{Rev} transcript, but with the additional specification of ``Other - African American" in the accent information, which we expected to recruit transcribers more familiar with the variety.
  \item \textbf{Amberscript}: A verbatim transcript from Amberscript\footnote{\url{https://www.amberscript.com/en/}}. We were helped by a salesperson who matched our audio with transcribers deemed well-suited.
\end{itemize}
\textbf{\textit{Machine Versions}}
\begin{itemize}
    \item \textbf{Rev ASR}: Generated using Rev.com's internal verbatim ASR model\footnote{\url{https://docs.rev.ai/api/asynchronous/}}, which is described in greater detail in \cite[Section 3.1]{fox2022improving}.
    \item OpenAI's \textbf{Whisper}: Generated using OpenAI's API\footnote{\url{https://platform.openai.com/docs/guides/speech-to-text}} to their large-v2 Whisper \cite{radford2023robust} model.
\end{itemize}

It is important to clarify that the \textbf{CORAAL} transcript versions were developed by a team of linguistic researchers; each file passed through multiple stages of transcription and editing where a researcher had access to the whole audio file. The \textbf{Rev}, \textbf{Rev (+AA tag)}, and \textbf{Amberscript} versions on the other hand were developed by professional transcribers who were only given a section of the audio to work with; each section could then have its quality verified and improved upon by a senior transcriber. 


\section{Methods}
\label{methods}
We used the open source 
\textit{fstalign}\footnote{\url{https://github.com/revdotcom/fstalign/}} with default settings to align transcript pairs to produce alignments of every permutation of transcript pairs \cite{del2021earnings}. 
We generated tests\footnote{\url{https://github.com/revdotcom/speech-datasets/tree/main/coraal-multi}} for our three transcription difference source hypotheses: \textit{morpho-syntactic}, \textit{reductions}, and \textit{verbatim}. 

Our \textit{morpho-syntactic} tests are based on the features enumerated in \cite{rickford1999features, spears2019rickford}.
We could not translate all the features into potential transcription difference tests. For example, 19p in \cite{spears2019rickford} refers to stressed ``stay," but the stressed and unstressed versions cannot be differentiated in written form. Another example is 20d in \cite{rickford1999features}, which describes the past participle form being used as the past tense. However, many common verbs have irregular participle or past forms (e.g. ``see"/``seen"/``saw" and ``run"/``run"/``ran"), making it difficult to algorithmically test for this alternation. Of the tests we were able to develop, some failed to capture any transcript differences. The \textit{morpho-syntactic} hypothesis ultimately consisted of 17 tests. 

The \textit{reductions} hypothesis consists of common contractions as well as a set of conventions used by CORAAL transcribers for reductions. The common contractions test checks for ``she'd/'s/'ve/'ll/'re/'t" contractions and their longer forms (e.g. ``she would/did/had," "she is/has," etc.). The CORAAL reduced form test checks for whether a substitution is made up of a full form and reduced form pairing listed in the table on pages 21-22 of the CORAAL user guide\footnote{\url{http://lingtools.uoregon.edu/coraal/userguide/}}. 

Finally, our \textit{verbatim} tests checked for filler deletions, filler substitution (e.g. transcript 1 has ``uh" while transcript 2 has ``um"), restart deletion or lack of restart indication (e.g. ``you-" vs ``you"), and repetition deletion.

These hypotheses are in order of most to least indicative of speaker characteristics. AAE feature erasure captured by \textit{morpho-syntactic} differences results in a more SAE-looking transcript which will potentially misrepresent the speech signal.
\textit{Reductions} are prevalent in both AAE and SAE, and they do not generally change the expression meaning, unlike some of the alternations caught by the morpho-syntactic tests. They do, however, have pragmatic consequences in that reduced forms are considered vernacular \cite{davydova2021role}; someone speaking in a more formal event, e.g. an interview or a trial, might prefer their reduced speech to be transcribed as the long forms. Finally, \textit{verbatim} differences do not change the speech content, and speakers are disfluent in every register, though the social context can impact how speakers are disfluent \cite{harrington2021style}. 



\section{Results} \label{results}

\subsection{Word error rates}


\begin{figure}[h]
    \centering
    \includegraphics[width=0.45\textwidth]{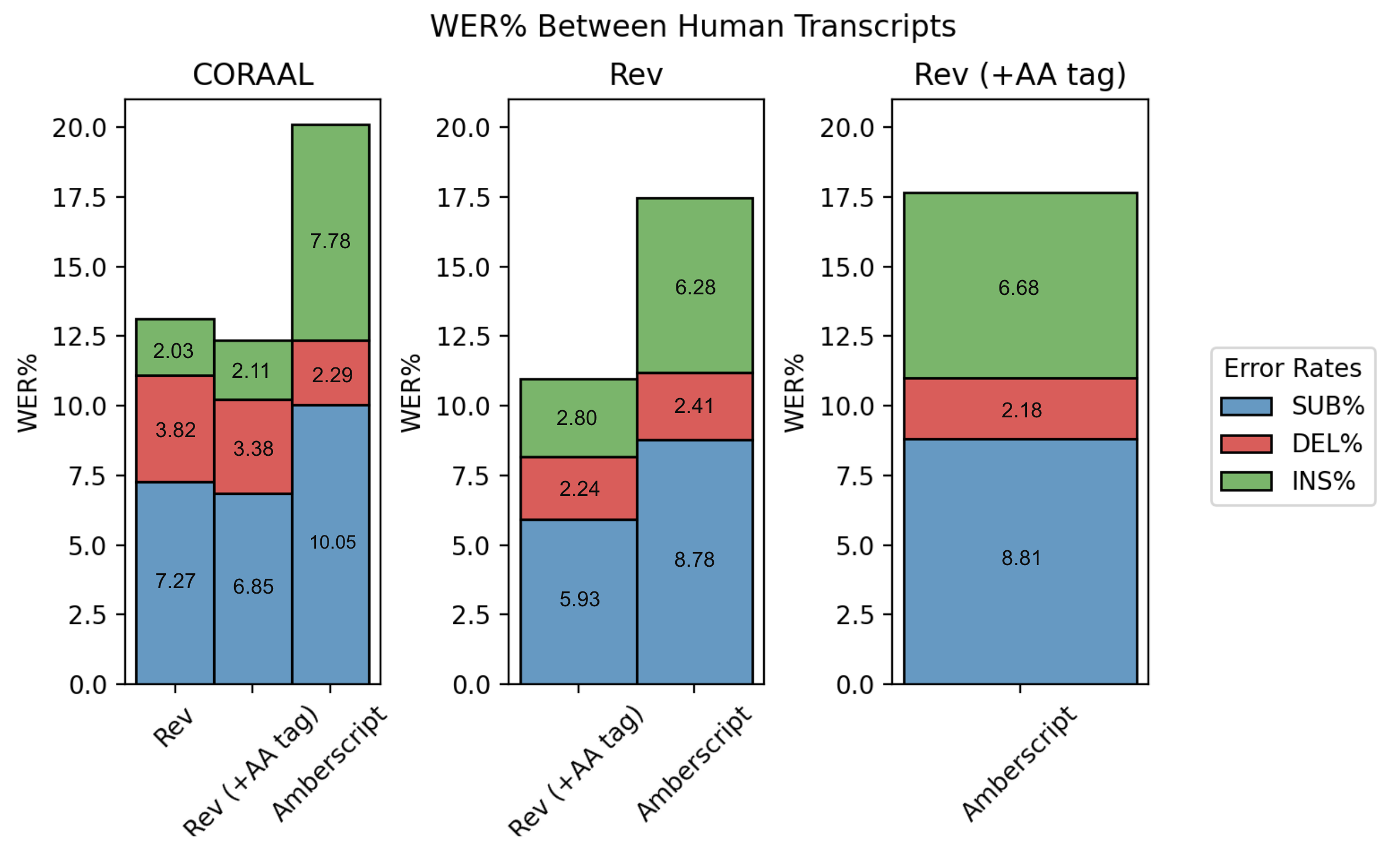}
    \caption{WER of human transcript pairs.}
    \label{refVariation}
\end{figure}

We first report WERs between the 4 human transcript versions. Though WER traditionally measures the error rate between an ASR hypothesis and a human reference, in this context we utilize the same WER mechanism to quantify the differences among humans by taking one human version as the reference and another as the hypothesis. We report the full WER as well as the individual rates of error that it is composed of, namely the rates of insertions (INS), deletions (DEL), and substitutions (SUB). As noted in \Cref{refVariation}, the WERs range between 10\% and 20\%, demonstrating the importance of the reference transcript to evaluation -- especially as many papers report traditional human vs machine WERs at much lower rates (e.g. \cite{radford2023robust,del2021earnings}). 

Unsurprisingly, the lowest WER is between the Rev and Rev (+AA tag) transcripts, likely because they were produced by a similar transcriber population. While the transcribers for the Rev (+AA tag) transcript may have been more familiar with AAE, they used the same style guide as the transcribers of the Rev transcript. It is even possible that there was overlap in the transcribers for the two sets of transcript versions. The Rev and Rev (+AA tag) versions also had relatively low WERs against the CORAAL transcript, suggesting similar stylistic proclivities. On the other hand, the greatest WER was between the CORAAL and Amberscript transcripts, most noticeably caused by the disproportionate amount of insertions. 

\begin{figure}[h]
    \centering
    \includegraphics[width=0.45\textwidth]{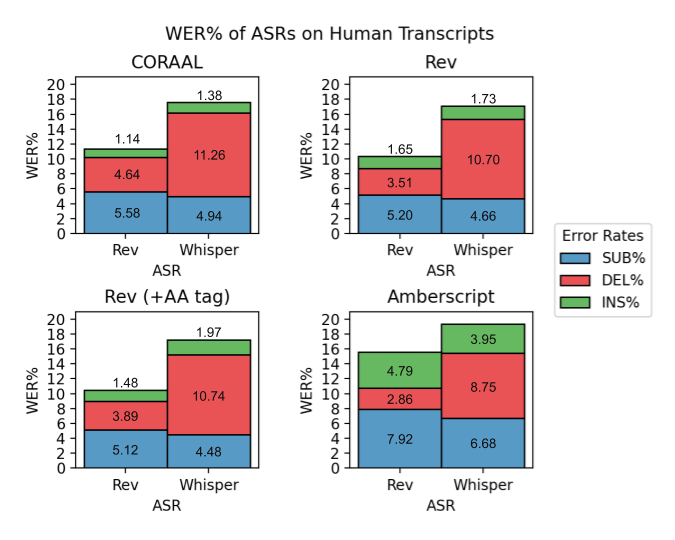}
    \caption{ASR WER against  human transcript versions.}
    \label{asrResults}
\end{figure}

We turn to the WERs of the Rev ASR and Whisper models, reported in \Cref{asrResults}. Rev's ASR performance is comparable between the CORAAL and both Rev transcript versions, but worse on Amberscript's version. Unexpectedly, we see a similar trend for the Whisper performance. We theorize that the higher deletion rate, compared to Rev, implies that the main difference between the models is likely where they fall on the verbatim to non-verbatim scale. We explore this hypothesis in the next section. 


\subsection{Difference source hypotheses}
\begin{figure}[h]
  \centering
  \includegraphics[width=\linewidth]{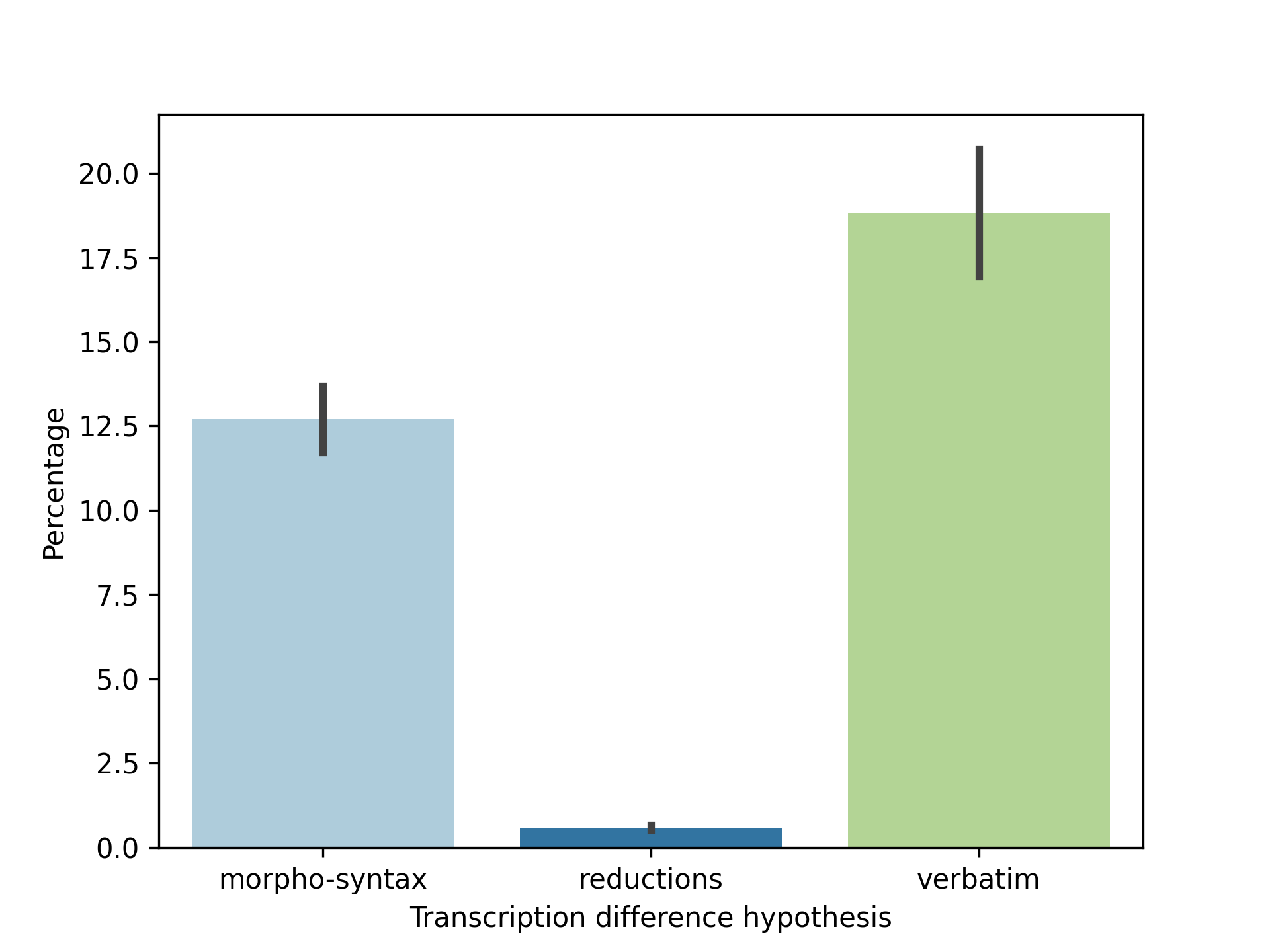}
  \caption{The average percentage of total differences that each hypothesis accounted for across each transcript pairwise comparison. The error bars correspond to the standard deviation across the transcript version pairwise comparisons.}
  \label{fig:BucketPercentages}
\end{figure}

Looking across all transcripts, \Cref{fig:BucketPercentages} shows that the biggest categories of differences are \textit{verbatim} and \textit{morpho-syntactic}, with \textit{reductions} accounting for very few differences. We tease out the impact of each of these two categories of differences per each transcript version pair.

\begin{figure}[h]
  \centering
  \includegraphics[width=\linewidth]{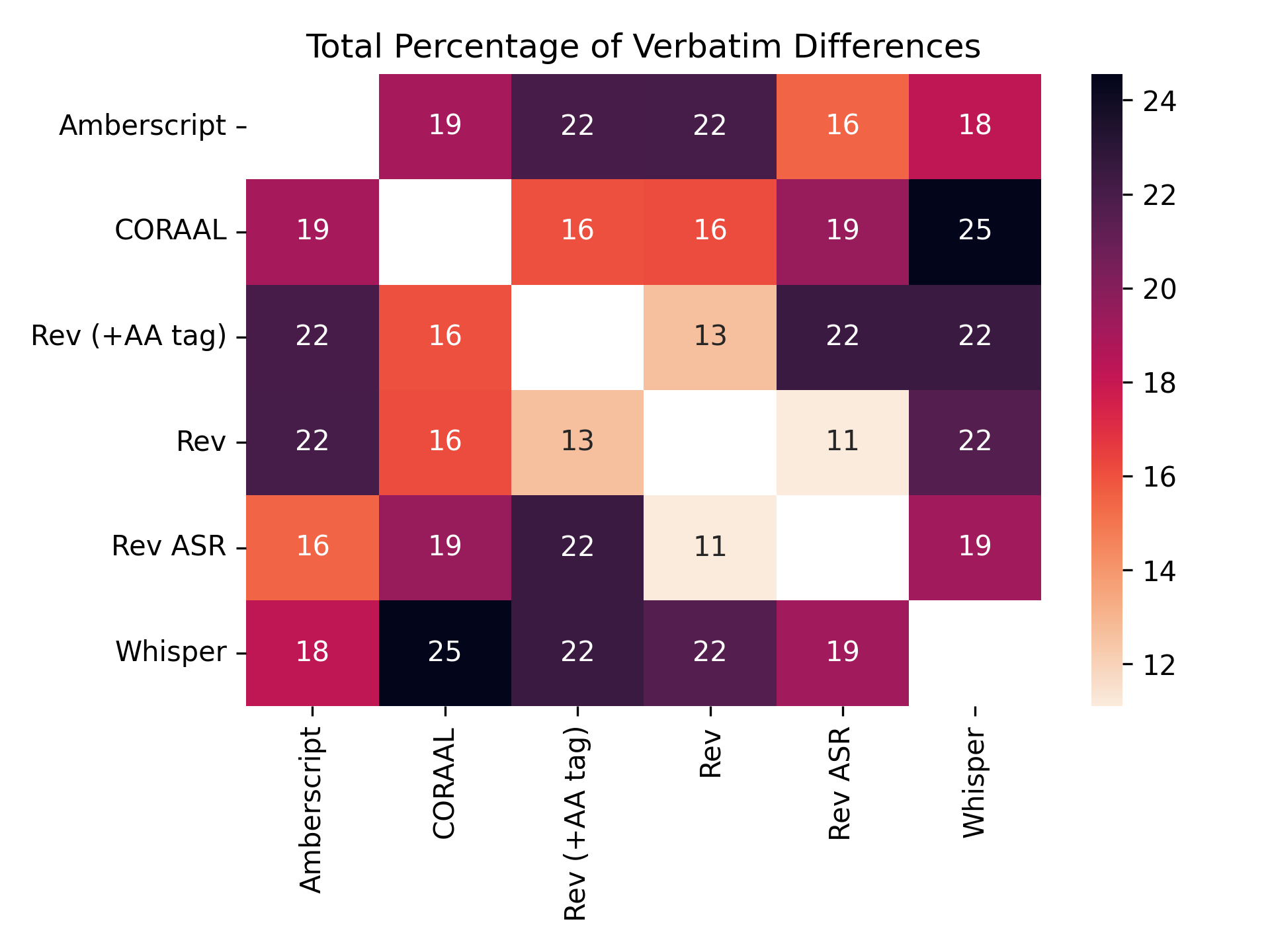}
  \caption{The percentage of differences between  transcript pairs for verbatim category.}
  \label{fig:Verbatim}
\end{figure}

\Cref{fig:Verbatim} verifies our hypothesis that a higher percentage of the differences between the Whisper transcript version and all the other versions are related to verbatim style choices. In fact, over all pairs, the greatest percentage of verbatim differences is between the Whisper and CORAAL versions while the lowest is between the Rev ASR model and the Rev versions. We note that the verbatim percentage between the Rev ASR model and the Rev (+AA tag) version is particularly large, larger than the difference between the two ASR models' transcript versions. The addition of the AA tag could have resulted in the transcribers taking greater liberty with respect to many parts of the style guide, including the verbatim instructions. 

\begin{figure}[t]
  \centering
  \includegraphics[width=\linewidth]{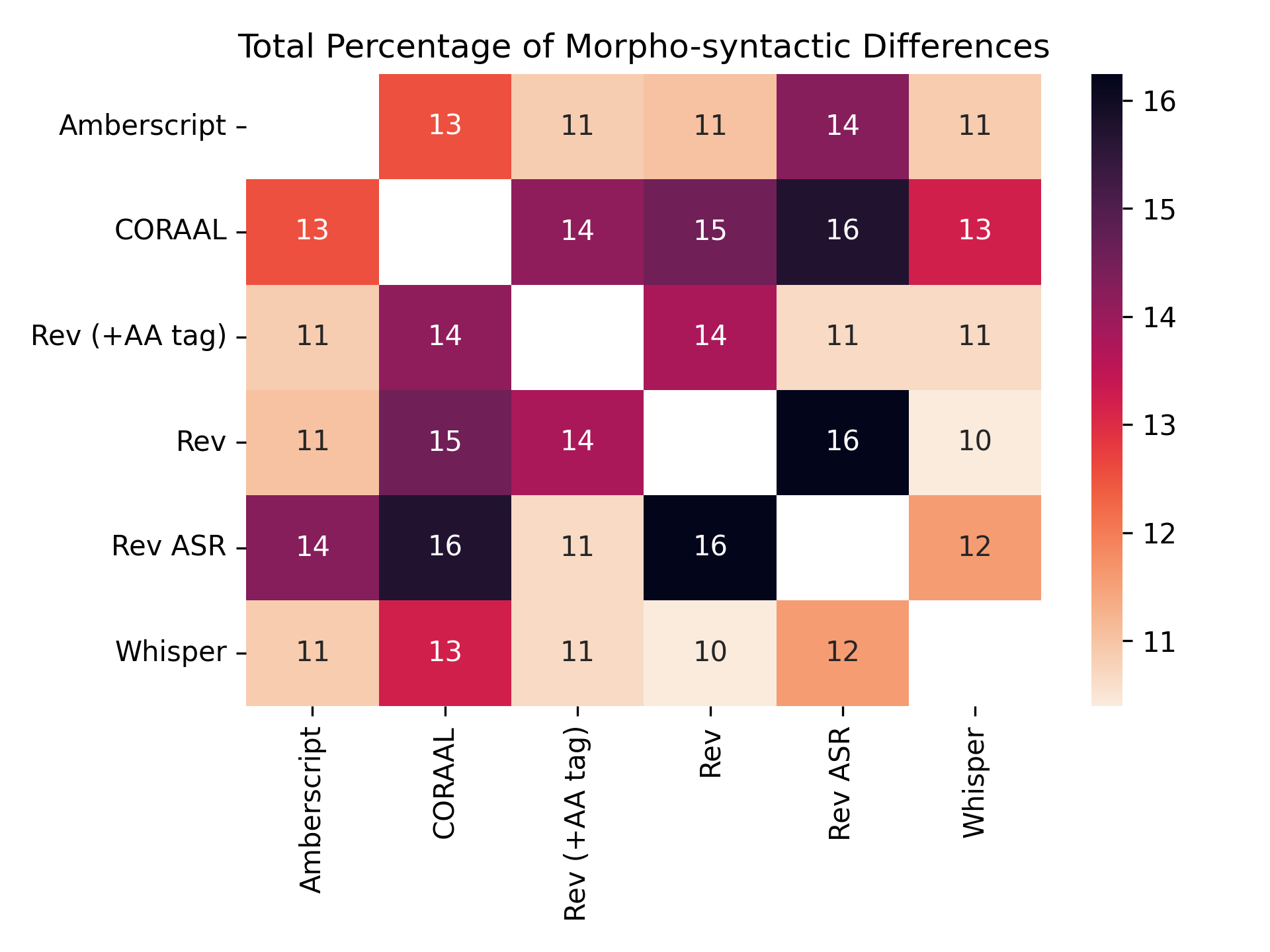}
  \caption{The percentage of differences between  transcript pairs for morpho-syntactic category.}
  \label{fig:Morphosyntactic}
\end{figure}

Looking into the morpho-syntactic differences, \Cref{fig:Morphosyntactic} shows that the Rev ASR vs Rev transcript versions and the Rev ASR vs CORAAL transcript versions have the highest percentage of these differences. In contrast, Rev's ASR transcript version vs the Rev (+AA tag) transcript percentage is relatively low. Particularly confusing is that the percentage of morpho-syntactic differences between the individual Rev (ASR and both human) versions is nearly the same as the percentage between each Rev version vs the CORAAL version. Because the CORAAL version was produced by linguists who are familiar with AAE and its morpho-syntactic features, we expect that the CORAAL transcript will typically be a more accurate representation of the speech in the audio. We believe that Rev transcribers and the ASR model may have more standardizing transcription proclivities that are causing these differences. We consider whether the style guide used by Rev transcribers could explain this in the following section.

\section{Discussion}

In this work, we investigated how and the extent to which reference transcripts of the same audio can differ, especially on underrepresented speech. We collected 6 transcript versions, 4 human- and 2 ASR-produced, of the same 10 hours of CORAAL. We found that the human-produced transcripts could vary by WERs as low as \(\sim \)10\% and as high as \(\sim \)20\%, and that ASR WER performance could increase or decrease by 5\% depending on the reference transcript. We also found the Rev human- and ASR-produced transcripts to be the most similar to one another. This makes sense because the transcribers were all trained on the same style guide and the ASR was trained on data from this same population of transcribers. We next examined three hypotheses about sources of stylistic differences, in order of most to least potentially misrepresentative: 1) morpho-syntactic differences between AAE and SAE, 2) reduction differences, and 3) verbatim vs non-verbatim differences. The verbatim hypothesis accounted for the greatest percentage of the transcript differences, and the morpho-syntatic hypothesis for the second most. The Rev transcripts for the most part had fewer verbatim differences than the other transcript version pairwise comparisons, but they interestingly had more morpho-syntactic differences. 

We might attribute this to the Rev style guide\footnote{\url{https://cf-public.rev.com/styleguide/transcription/Transcription+Style+Guide+v5.pdf}}, which instructs transcribers to ``use English grammar conventions while maintaining the integrity of what was spoken. We are unable to cover and address specific guidelines regarding grammar. We expect you to have prior knowledge of, or be able to research, American English grammar, capitalization, and punctuation guidelines." This is ambiguous with respect to non-standard language varieties. Many AAE features that we tested for are often taught to be ``ungrammatical" in schools \cite{fogel2006teaching}. At the same time, those more familiar with AAE might deem them necessary to ``maintaining the integrity of what was spoken." Rev might consider clarifying this part of the style guide for underrepresented language varieties, as well as augmenting the customer-facing definition of verbatim vs non-verbatim, or introducing a new transcription variety option. The inclusion of examples could help as well. Then a user wanting to have audio of a non-standard variety transcribed could choose whether their variety's morpho-syntactic features are transcribed with the standard variety's constructions or not (\cite{miller-etal-2018-embedding} makes a similar proposal for machine translation). Of course, greater awareness and education about underrepresented varieties would also help with this.



With this work, we add to the ever more important research into bias in machine learning. We give more insight to similar discrepancies found by \cite{koenecke2020racial} and identify key categories of errors. Moreover, we come to the conclusion that a single reference transcript may not be sufficient to conclusively make claims about performance. Our findings indicate that different transcript versions may highlight distinct, yet equally valid, variations (e.g. verbatim vs non-verbatim) that must be considered for fair evaluation. We hope that by making our transcript versions and code available, we assist other research in addressing the important impact of human variation and bias.

\bibliographystyle{IEEEtran}
\bibliography{mybib}

\end{document}